\documentclass{article}
\usepackage{spconf,amsmath,graphicx,hyperref}

\usepackage{cite}
\usepackage{amsmath,amssymb,amsfonts}
\usepackage{algorithmic}
\usepackage{graphicx}
\usepackage{textcomp}
\usepackage{xcolor}
\usepackage{amssymb}
\usepackage[capitalize]{cleveref}
\usepackage{color}
\usepackage{booktabs} 
\usepackage{url}
\usepackage{algorithm}
\usepackage{algorithmic}
\usepackage{multirow}
\usepackage{setspace}

\begin{document}

\title{Parametric Neural Amp Modeling with Active Learning}
\name{Florian Grötschla${^*}$ \qquad Longxiang Jiao${^*}$ \qquad Luca A. Lanzendörfer \qquad Roger Wattenhofer
\thanks{$^*$Equal contribution}}
\address{ETH Zurich}

\newcommand{\model}{\textsc{PANAMA}}

\maketitle

\begin{abstract}
We introduce \model, an active learning framework to train parametric guitar amp models end-to-end using a combination of an LSTM model and a WaveNet-like architecture. With \model, one can create a virtual amp by recording samples that are determined through an ensemble-based active learning strategy to minimize the amount of datapoints needed (i.e., amp knob settings). Our strategy uses gradient-based optimization to maximize the disagreement among ensemble models, in order to identify the most informative datapoints. MUSHRA listening tests reveal that, with 75 datapoints, our models are able to match the perceptual quality of NAM, the leading open-source non-parametric amp modeler.
\end{abstract}


\section{Introduction}\label{sec:introduction}
In recent years, data-driven guitar amp modeling has become increasingly popular. Such approaches treat an amp as a blackbox, and simply learn the transform which the amp applies to the raw guitar signal in an end-to-end fashion. The trained models can then be integrated as plugins into Digital Audio Workstations (DAWs), or deployed on modeler pedals. In general, one can distinguish between two types of such amp models: Parametric models, which are conditioned on the amp settings, allowing virtual knobs to be adjusted freely during inference; and non-parametric models, also known as captures, which aim to create a snapshot of the amp with the knobs at a particular setting.     
Most existing blackbox amp models use either feedforward variants of WaveNet~\cite{damskägg2019wavenet} or RNNs, frequently  using LSTM cells~\cite{wright2019RNN}. Both have demonstrated sufficient accuracy and low latency for real-time inference on consumer hardware~\cite{wright2020realtime}. Among open-source efforts, Neural Amp Modeler (NAM)~\cite{NAM} is the leading solution. It allows users to capture their own amp setups and supports both WaveNet and LSTM architectures. However, unlike commercial products such as NeuralDSP%
or AmpliTube
, NAM does not offer parametric knob controls. GuitarML
is another open-source amp modeler, which supports conditioning models on a single control parameter, but does not extend this functionality to multiple knobs.
To our knowledge, there are currently no open-source parametric amp modelers allowing a full range of virtual knob controls. The biggest challenge hindering the practicality of such modelers is arguably the cumbersome data collection procedure. This involves turning the amp knobs to numerous configurations and recording the amp responses, where the number of required measurements grows exponentially with the number of knobs. In this work, we aim to address this issue using active learning.  
Our main contributions in this paper are: 1) \model{} (\textbf{PA}rametric \textbf{N}eural \textbf{A}mp \textbf{M}odeling with \textbf{A}ctive learning), an open-source parametric amp modeler; 2) An active learning training framework which determines the optimal datapoints to be collected and thus minimizes the total amount. To this end, we make use of model ensembling and introduce a method to find the most informative datapoints in a continuous space using gradient-based optimization.

\section{Related Work}

\subsection{Guitar Amp Modeling}
Although there exist white-box approaches for the modeling of audio effects~\cite{Esqueda2021DifferentiableWV}, which aim to explicitly simulate the internal workings of amplifier components, these methods often require extensive computational resources and detailed knowledge of physical parameters, making them impractical. In this paper, we are concerned with black-box models, which focus on learning the complex relationship between input and output directly from data.
Damskägg et al. were the first to use a WaveNet feedforward variant for the amp modeling task~\cite{damskägg2019wavenet}. Their approach outperformed both a baseline MLP model and a block-oriented model, and they also showed that a C++ implementation could run real-time on a consumer-grade computer~\cite{Damskgg2019RealTimeMO}.
RNNs were first applied to amp modeling by Covert and Livingston with a NARX network~\cite{Covert2013AVG}. Zhang et al. later experimented with LSTMs~\cite{Zhang2018AVG}, though both studies reported limited accuracy. Schmitz and Embrechts subsequently demonstrated more successful LSTM-based approaches, validated through listening tests~\cite{Schmitz2019NonlinearMO}.
Wright et al. conducted a comparison between LSTM, GRU, and WaveNet models for both amp and pedal modeling~\cite{wright2019RNN}, finding that WaveNet performed best for amps while LSTM was more accurate for pedals. A follow-up paper confirmed both real-time feasibility and the superiority of LSTM over GRU~\cite{wright2020realtime}. LSTM models remain in use in recent work, including by NeuralDSP~\cite{juvela2024endtoendampmodelingdata}, while feedforward WaveNet is the default architecture in NAM~\cite{NAM}.
Beyond amps, these architectures have been widely used for modeling nonlinear audio effects. WaveNet and RNNs have proven effective for distortion pedals~\cite{Yoshimoto2021WaveNetMO, Sudholt2023PruningDN}, while LSTMs have been applied to time-varying effects such as phasers and flangers~\cite{Wright2021NeuralMO}. Steinmetz and Reiss proposed efficiency-optimized TCNs for compressors~\cite{Steinmetz2021EfficientNN}, and Ramirez et al. introduced encoder-decoder and latent-space architectures combining RNN and WaveNet components~\cite{Ramirez2020latentWavenet}. Comunita et al. further incorporated time-varying modulation to better capture long-range dependencies~\cite{Comunit2022ModellingBA}.
The only prior work to address the difficult data collection procedure for parametric amp modeling is by NeuralDSP~\cite{juvela2024endtoendampmodelingdata}, who used a robotic system to automate knob turning and recording. However, such an approach is impractical for end users. Our work is the first to introduce active learning to address this problem.

\subsection{Active Learning}
Active learning is a training paradigm that reduces labeling costs by selectively querying informative datapoints. Rather than training on a randomly sampled dataset, the model iteratively selects datapoints based on an acquisition function that estimates their potential to improve performance.
In early work, Seung et al. introduced Query-by-Committee~\cite{Seung1992QueryBC}, which selects datapoints that maximize model disagreement. Ensemble-based methods remain influential in modern deep learning~\cite{lakshminarayanan2017deepensembles}.
Most existing active learning approaches assume a pool-based setting~\cite{zhan2021poolal}, where an unlabeled dataset exists and models query from it. By contrast, our method operates without a pool, directly searching for optimal amp settings to query in a continuous space. This corresponds to Membership Query Synthesis, introduced by Angluin~\cite{angluin1988}.

\section{Methodology}

\begin{figure}
    \centering
    \includegraphics[width=0.5\textwidth]{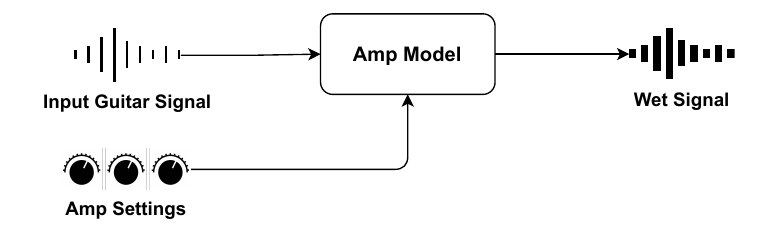}
    \caption{Single Model Setup. The parametric amp model transforms the DI guitar input signal, conditioned on the settings of virtual knobs.}
    \label{fig:basic}
\end{figure}

\subsection{Model Architecture} 
The goal of our model is to transform the input signal conditioned on amp settings, as visualized in \cref{fig:basic}. For the main architecture, we adapt WaveNet~\cite{oord2016wavenetgenerativemodelraw} to operate in a feed-forward manner, similar to prior works. WaveNet was originally developed as an autoregressive generative model, which processes audio using a stack of dilated convolutional layers and predicts a categorical distribution over the next sample in sequence. Here, we use these convolutional layers to transform the input signal directly. 
Let $\mathbf{x}$ be the input guitar signal, $\mathbf{g}$ be a vector containing the value of each knob on the amp as a real number in $[0, 1]$. The original WaveNet architecture provides a way to add both local and global conditioning to the model, which we adopt here. The local condition is a time series $\mathbf{c}$, where each timestep is designed to affect the corresponding timestep from the input. Following NAM~\cite{NAM}, we set $\mathbf{c} = \mathbf{x}$. The global condition, on the other hand, is a single static vector designed to have effect across all timesteps, which we set to $\mathbf{g}$. Both conditions are incorporated into the convolutional layers as follows: 
\[
\mathbf{z} = \tanh(W_{f} * \mathbf{x} + V_{f} * \mathbf{c} + V_{f}'^T \mathbf{g}) \odot \sigma(W_{g} * \mathbf{x} + V_{g} * \mathbf{c} + V_{g}'^T \mathbf{g})
\]
where $W_{*}$ are convolutional kernels, $V_{*}$ are 1x1 convolutional kernels, and $V'_{*}$ are linear mappings. The vector $V'^{T}_{*}\mathbf{g}$ is broadcast over the time dimension. %
Additionally, we employ a RNN model using LSTM. Here, we simply broadcast the amp settings $\mathbf{g}$ over the time dimension and concatenate it onto the input $\mathbf{x}$. At timestep $t$, updating the hidden state $\mathbf{h}$ and deriving the model output $\mathbf{y}$ can be expressed as:
\[
\mathbf{h}_t = f(\operatorname{cat}(\mathbf{x}_t, \mathbf{g}), \mathbf{h}_{t-1}),\ \mathbf{y}_t = g(\mathbf{h}_t)
\]

\begin{figure}
    \centering
    \includegraphics[width=0.5\textwidth]{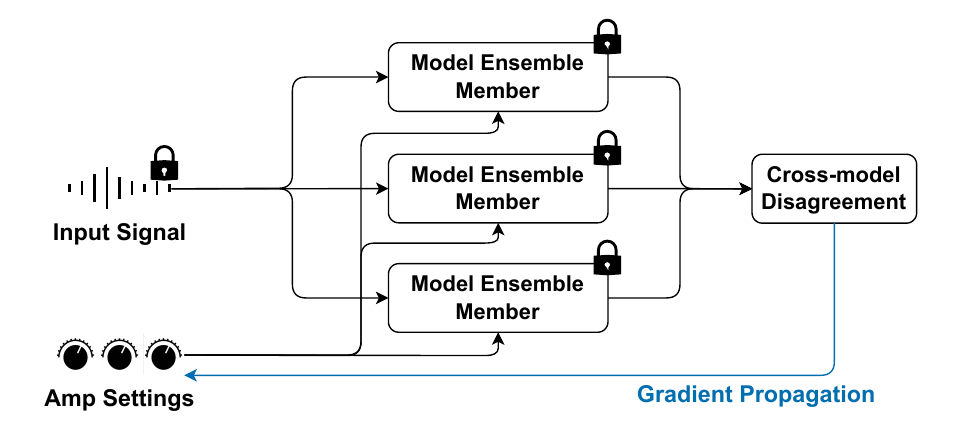}
    \caption{Active Learning Setup. The input signal $\mathbf{x}$ and amp settings $\mathbf{g}$ are fed into independently trained instances of the model, and variation in the outputs is calculated as the cross-model disagreement $D$. Gradients are propagated back to $\mathbf{g}$ in order to maximize $D$.}
    \label{fig:actlearn}
\end{figure}
 \begin{figure}[b]
     \centering
     \includegraphics[width=0.45\textwidth]{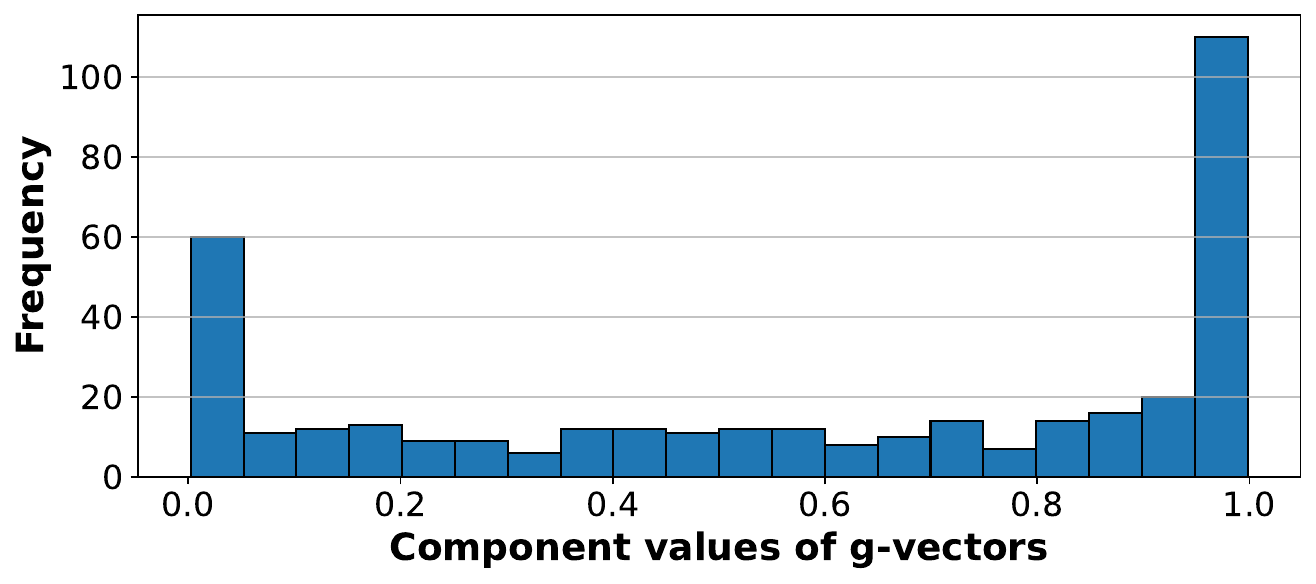}
     \caption{Distribution of component values in $\mathbf{g}$-vectors gathered through active learning.}
     \label{fig:g_dist}
 \end{figure}

\subsection{Active Learning Approach}
In this paper, we take the following perspective: An input signal $\mathbf{x}$ and an amp settings vector $\mathbf{g}$ together constitute an unlabeled datapoint, whereas their corresponding wet signal $\mathbf{y}$ from the amp is considered the ground truth label. We keep $\mathbf{x}$ as a pre-chosen, fixed signal and vary $\mathbf{g}$. Then, a dataset consists of datapoint-label pairs in the form of $\{((\mathbf{x}, \mathbf{g^{(i)}}), \mathbf{y^{(i)}})\}_{i=1}^N$. The goal of our active learning method is to find the optimal next datapoints to ``label'' (which involves running the fixed input signal through the amp with the given settings, and recording the output audio), given the current set of already labeled points, denoted $\mathcal{L}$. 
While predictive uncertainty would be the theoretically ideal acquisition function, our deterministic model does not provide uncertainty estimates directly. Instead, we use the output disagreement among members of an LSTM ensemble, denoted $D$, as a proxy, inspired by Query-by-Committee~\cite{Seung1992QueryBC}. Intuitively, strong disagreement among ensemble models over the predicted output signal of a datapoint indicates high uncertainty in that region of the input space, and thus suggests that the datapoint is likely to be informative if labeled. 
Suppose that we have a model ensemble $\mathcal{F}_\mathcal{L}$ of size $M$, where $f^{(i)}_\mathcal{L} \in \mathcal{F}_\mathcal{L}$ denotes the $i$-th model. Each $f^{(i)}_\mathcal{L}$ is trained independently with the current labeled dataset $\mathcal{L}$, using random initialization, dataset shuffling etc. so that they converge to slightly different results. Then, the disagreement of $\mathcal{F}_\mathcal{L}$, denoted $D_\mathcal{L}$, can be calculated as:
\begin{align}
D_\mathcal{L}(\mathbf{x}, \mathbf{g}) &= w_{\text{waveform}} \frac{1}{M} 
    \operatorname{tr}\!\Big(\operatorname{Var}_i[f^{(i)}_\mathcal{L}(\mathbf{x}, \mathbf{g})]\Big) \nonumber\\
    &\quad + w_{\text{mel}} \frac{1}{M} 
    \operatorname{tr}\!\Big(\operatorname{Var}_i[\operatorname{Mel}(f^{(i)}_\mathcal{L}(\mathbf{x}, \mathbf{g}))]\Big) \nonumber
\end{align}
where $\operatorname{Var}$ denotes the covariance matrix, $\operatorname{Mel}$ computes the (vectorized) Mel-spectrogram, and $w_*$ are scalar weights. In other words, for both the waveform and Mel-spectrogram domains, we measure the element-wise variances of the output across ensemble members, and aggregate them into a scalar. As $\mathbf{x}$ is fixed, we optimize over $\mathbf{g}$, defining the optimal $\mathbf{g^{*}}$ as:
\[
\mathbf{g^{*}} := \operatorname{argmax}_{\mathbf{g} \in \mathbb{R}^k; 0 \leq \mathbf{g} \leq 1} D_\mathcal{L}(\mathbf{x, g}),
\]
where $k$ is the number of settings, or knobs on the amp to be modeled.
Importantly, the disagreement $D_\mathcal{L}$ is differentiable w.r.t. the output signals $f^{(i)}_\mathcal{L}$, which are again differentiable w.r.t. $\mathbf{g}$. Thus, we can reprogram the gradient from $D_\mathcal{L}$ to $\mathbf{g}$, allowing us to use any gradient-based optimization algorithm to find $\mathbf{g^{*}}$. An overview of the approach can be found in \cref{fig:actlearn}.
We restart the optimization process several times, collecting the distinct optima obtained in a batch $\mathbf{G}$. Subsequently, $\mathbf{x}$ is run through the amp using each setting $\mathbf{g^{(i)}} \in \mathbf{G}$, and the output $\mathbf{y^{(i)}}$ is recorded. The datapoint $((\mathbf{x}, \mathbf{g^{(i)}}), \mathbf{y^{(i)}})$ is then added to $\mathcal{L}$. 
The process of training the ensemble, finding the optimal amp settings, recording the amp signals, and expanding the dataset is repeated iteratively. After enough datapoints are collected, we train a final model using WaveNet feedforward. The pseudocode describing this process is shown in \cref{pcode}. 

\begin{algorithm} 
\caption{Active Learning} 
\begin{algorithmic}[1]
\STATE $\mathbf{Amp} \leftarrow$ Guitar amp to be modeled.
\STATE $\mathbf{x} \leftarrow$ Fixed training signal.
\STATE $\mathcal{L} \leftarrow$ Initial labeled dataset.
\FOR{$t=1$ to $T$}
    \STATE Train ensemble $\mathcal{F}_\mathcal{L} = \{f^{(i)}_{\mathcal{L}}\}_{i=1}^{M}$ on $\mathcal{L}$.
    \STATE $D_\mathcal{L}(\mathbf{x}, \mathbf{g}) \leftarrow \text{Variance among } f^{(i)}_\mathcal{L}(\mathbf{x}, \mathbf{g})$
    \STATE $\mathbf{G} \leftarrow \operatorname{Optimize}_{\mathbf{g}} D_\mathcal{L}(\mathbf{x}, \mathbf{g})$. %
    \STATE $\mathcal{L} \leftarrow \mathcal{L} \cup \{((\mathbf{x}, \mathbf{g}), \mathbf{y}) \mid \mathbf{y} = \operatorname{Amp}(\mathbf{x}, \mathbf{g}), \mathbf{g} \in \mathbf{G}\}$.
\ENDFOR
\STATE Train final model $f$ on $\mathcal{L}$.
\RETURN $f$.
\end{algorithmic}
\label{pcode}
\end{algorithm}

\section{Experimental Evaluation}
For our experiments, we choose $\operatorname{dim}(\mathbf{g}) = 6$. This includes the following parameters: Gain, Bass, Mid, Treble, Master, Presence. The Master knob is included, as its behavior differs from a standard volume knob on tube amps. The fixed input signal is taken from NAM~\cite{NAM} and roughly 3 minutes long. We obtain the ground truth signals from an amp sim. Throughout the experiments, we use a weighted combination of MSE and multi-scale mel-spectrogram loss. The settings for the latter are adopted from Descript Audio Codec~\cite{kumar2023highfidelityaudiocompressionimproved}. For testing, we use 30 minutes of guitar audio across different genres, gathered from the IDMT-SMT-GUITAR dataset~\cite{kehling_2023_idmt_smt_guitar}, and around 1,000 randomly sampled amp settings. Code is available online~\footnote{\scriptsize{\url{https://github.com/ETH-DISCO/PANAMA}}}.

\begin{figure}
    \centering
    \includegraphics[width=\linewidth]{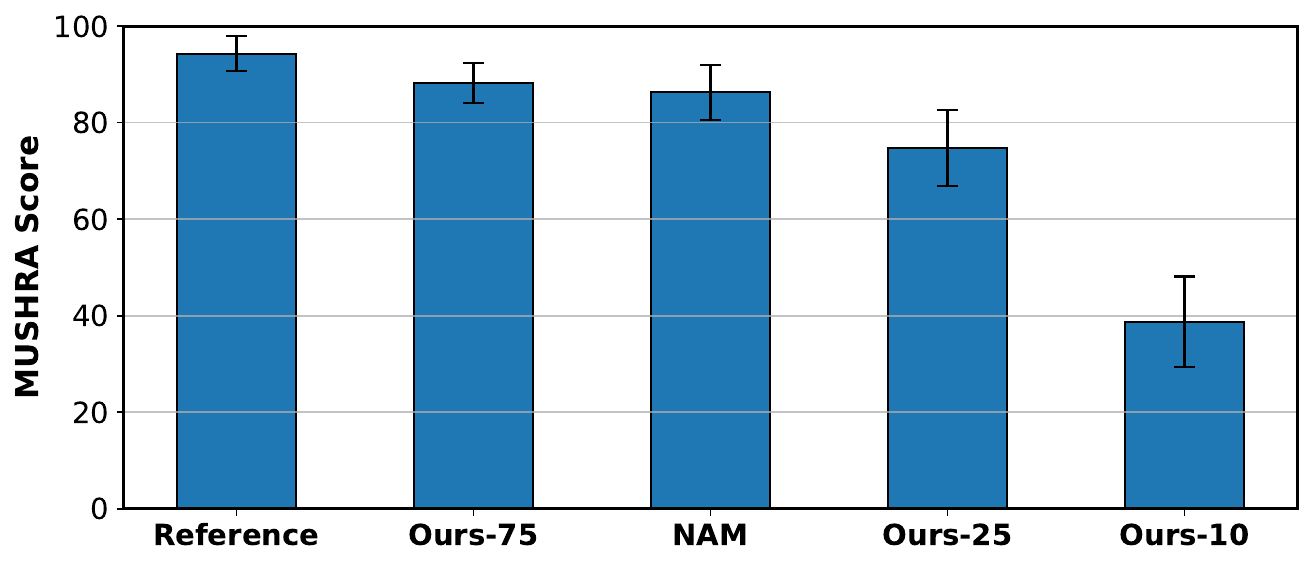}
    \caption{Comparison of MUSHRA scores across models. Bars show the mean score and error bars represent the 95\% confidence interval. ``Ours-10/25/75” indicate our models trained with 10, 25, and 75 active learning samples, respectively. Ours-75 matches the quality of NAM, the leading open-source non-parametric amp modeler.}
    \label{fig:mushra}
\end{figure}
\subsection{Active Learning} 
We use a model ensemble of size 4 and initialize a starting dataset with 10 randomly sampled, labeled points. In each round, each model in the ensemble is first trained on the currently gathered datapoints. Then, an Adam optimizer is used to maximize the disagreement among ensemble members 10 independent times with respect to $\mathbf{g}$. From the resulting 10 candidate optima, we extract the unique ones using a clustering algorithm. This usually yields 6-7 distinct $\mathbf{g}$-vectors per round. 
We retrain the final model after 0, 2 and 10 rounds of active learning, corresponding to datasets of size 10, 25 and 75 respectively. To evaluate performance, we conduct a MUSHRA listening test comparing the three models with NAM. 7 different clips from the IDMT-SMT-GUITAR dataset~\cite{kehling_2023_idmt_smt_guitar} are selected for the test, 2 used to familiarize participants with the setup, and 5 used for the actual evaluation. Different amp settings are chosen to reflect common real-world use. For NAM, which is non-parametric, a separate model is trained for each amp setting, whereas our models are conditioned on the different settings during inference. 
After filtering out individuals who rated the reference too low, 10 participants remain for the final analysis. Results are shown in \cref{fig:mushra}.
The MUSHRA scores reveal no significant difference between our model trained with 75 samples (Ours-75) and NAM, indicating that our parametric model is able to match the perceptual quality of non-parametric NAM captures. Moreover, both methods achieve performance that is comparable to that of the reference.

\subsection{Ablation against random and heuristic sampling}
We compare three strategies for selecting training points: (i) our active learning method, (ii) random sampling, and (iii) a heuristic sampling scheme inspired by the behavior of active learning.
For (ii), we sample datapoints from a uniform random distribution. (iii) is motivated by our observation that the vectors chosen by active learning tend to have component values near 0 or 1. 
\begin{table}[t]
    \centering
    \resizebox{\columnwidth}{!}{
        \begin{tabular}{lcccc}
            \toprule
            \textbf{Ensemble} & \textbf{Final} & \textbf{Ens. Train. Speed} & \textbf{Test MSE}$\downarrow$ & \textbf{Test Mel}$\downarrow$\\
            \midrule
            LSTM & LSTM & 14.6M samples/s & 3.06e-04 & 3.72 \\ 
            WaveNet & WaveNet & 3.1M samples/s & 2.54e-04 & 3.17 \\
            LSTM & WaveNet & 14.6M samples/s & \textbf{1.61e-04} & \textbf{2.55} \\ 
            \bottomrule
        \end{tabular}
    } 
    \caption{Comparison of different architectural combinations in active learning to sample 75 datapoints. Combining LSTM and WaveNet outperforms using either architecture alone, while inheriting the fast training speed of LSTM.}
    \label{tab:arch_comparison}
\end{table}
This is visualized in \cref{fig:g_dist}, where the gathered vectors are flattened into a single list of scalars and plotted as a histogram. 
Knob settings close to the min/max values should be more challenging for the model to interpolate from previously seen settings. At the same time, it raises the question of whether heuristically sampling such extreme values would already suffice. The histogram resembles the shape of a beta distribution, and fitting one yields parameters $\alpha = 0.5396$ and $\beta = 0.4122$. Based on this, we design (iii) to sample datapoints directly from a beta distribution with $\alpha = \beta = 0.5$.
Each method is used to collect a total of 75 datapoints. Once this limit is reached, a WaveNet model is trained on the resulting dataset and evaluated. %
The beta heuristic performs worse than random sampling, with a test MSE of 5.80e-04 vs 3.16e-04 for random sampling and mel losses on the test set of 5.06 vs 3.49. This indicates that naive strategies biased toward extreme values can actually harm performance. In contrast, our active learning approach achieves substantially lower validation errors with a test MSE of 1.61e-04 and a Mel of 2.55, producing more accurate models under the same data budget. These results suggest that although active learning favors extreme values, its strategy is considerably more sophisticated than the simple heuristic.

\subsection{Ablation on the effect of different architectures}
We evaluate different architectural combinations during ensemble and final model training, as shown in \cref{tab:arch_comparison}. The training speed is measured as the number of audio samples per second on RTX 3090 GPUs. 
The combination of LSTM and WaveNet outperforms the two single-architecture approaches, achieving both the highest ensemble training speed and the best performance. This suggests that LSTM is faster and more effective at selecting informative datapoints during the acquisition stage, whereas WaveNet is better at exploiting the datapoints to produce high quality final models. Combining them leverages the strengths of both.

\section{Conclusion}
We introduced \model, a parametric guitar amp modeling framework using active learning to minimize the number of datapoints needed. We made use of an LSTM model ensemble to calculate the disagreement as the acquisition function, and propagated the gradient back to the amp settings, allowing us to run gradient-based optimization to find the most informative settings in a continuous space. MUSHRA listening tests showed that, with 75 datapoints gathered this way, a feedforward WaveNet model trained on the final dataset matches the quality of the leading open-source non-parametric amp modeler NAM. 

\bibliographystyle{IEEEtran}
\bibliography{ISMIRtemplate}

\end{document}